\documentclass{article}
\usepackage{amsmath, amsfonts, amssymb,amsthm,multicol,derivative,enumitem,graphicx,float,tikz,tikz-qtree,forest,verbatim,bm,spconf, lipsum,afterpage}
\usepackage[utf8]{inputenc}
\usepackage[mathscr]{euscript}
\usepackage[ruled,linesnumbered]{algorithm2e}
\usetikzlibrary{fit,bayesnet, positioning, quotes,calc}
\usepackage[numbers,sort&compress]{natbib}
\usepackage{hyperref}

\title{A PLS-Integrated LASSO Method with Application in Index Tracking}
\name{Shiqin Tang\textsuperscript{1},\, Yining Dong\textsuperscript{1},\, S. Joe Qin\textsuperscript{2}}
\address{\textsuperscript{1}School of Data Science, City University of Hong Kong, Hong Kong\\ \textsuperscript{2}Department of Computing and Decision Sciences, Lingnan University, Hong Kong\\
\href{mailto:t.sq@my.cityu.edu.hk}{\nolinkurl{t.sq@my.cityu.edu.hk}}, \href{mailto:yining.dong@cityu.edu.hk}{\nolinkurl{yining.dong@cityu.edu.hk}}, \href{mailto:joeqin@ln.edu.hk}{\nolinkurl{joeqin@ln.edu.hk}}
} 

\begin{document}
\maketitle

% Math formats
\theoremstyle{defn}
\newtheorem*{defn}{Definition}
\theoremstyle{thm}
\newtheorem*{thm}{Theorem}
\theoremstyle{lem}
\newtheorem*{lem}{Lemma}
\theoremstyle{rmrk}
\newtheorem*{rmrk}{Remark}
\theoremstyle{rslt}
\newtheorem{rslt}{Result}[section]
\theoremstyle{expl}
\newtheorem*{expl}{Example}
\theoremstyle{cor}
\newtheorem*{cor}{Corollary}
\theoremstyle{smry}
\newtheorem*{smry}{Summary}

% Math Command
\let\oldforall\forall
\renewcommand{\forall}{\oldforall \, }
\let\oldLeftrightarrow\Leftrightarrow
\renewcommand{\Leftrightarrow}{\oldLeftrightarrow \,\, }
\let\oldexist\exists
\renewcommand{\exists}{\oldexist \: }
\newcommand\existu{\oldexist! \: }
\newcommand{\st}{\mathrm{\,\,s.t.\,\,}}

% Math functions
\newcommand{\KL}{D_{\mathrm{KL}}}
\newcommand{\proj}{\mathcal{P}}
\let\oldvec\vec 
\renewcommand{\vec}{\mathrm{vec}}
\newcommand{\diag}{\mathrm{diag}}
\newcommand{\tr}{\mathrm{tr}}
\newcommand{\indep}{\perp \!\!\! \perp}
\newcommand{\notindep}{\not\!\perp\!\!\!\perp}

% stats & probability
\newcommand{\Normal}{\mathcal{N}}
\newcommand{\Ga}{\mathrm{Ga}}
\newcommand{\IG}{\mathrm{IG}}
\newcommand{\Var}{\mathrm{Var}} % var
\newcommand{\V}{\mathbb{V}} % var
\newcommand{\I}{\mathbb{I}} % Fisher info
\newcommand{\cL}{\mathcal{L}} % Lagrangian & ELBO (curly L)
\newcommand{\E}{\mathbb{E}}
\newcommand{\xhat}{\hat{\mathbf{x}}}
\newcommand{\phat}{\hat{p}}
\newcommand{\ptil}{\tilde{p}}
\newcommand{\ftil}{\tilde{f}}
\newcommand{\xbar}{\bar{\mathbf{x}}}
\newcommand{\ybar}{\bar{\mathbf{y}}}
\newcommand{\thetahat}{\hat{\theta}}
\newcommand{\wmle}{\hat{\mathbf{w}}_{\mathrm{MLE}}}
\newcommand{\what}{\hat{\mathbf{w}}}
\newcommand{\muhat}{\hat{\mu}}
\newcommand{\thetatil}{\tilde{\theta}}
\newcommand{\mutil}{\tilde{\mu}}
\newcommand{\tautil}{\tilde{\tau}}
\newcommand{\phibar}{\bar{\phi}}
\newcommand{\Phibar}{\bar{\Phi}}
\newcommand{\Abar}{\bar{A}}
\newcommand{\etabar}{\bar{\eta}}
\newcommand{\xihat}{\hat{\xi}}
\newcommand{\Ntil}{\tilde{\mathcal{N}}}
\newcommand{\ELBO}{\mathrm{ELBO}}

% Bold-face symbols & curly symbols
\newcommand{\bfphi}{\mathbf{\Phi}}
\newcommand{\bfw}{\mathbf{w}}
\newcommand{\bfx}{\mathbf{x}}
\newcommand{\bfz}{\mathbf{z}}
\newcommand{\bfy}{\mathbf{y}}
\newcommand{\bfalpha}{{\bm{\alpha}}}
\newcommand{\bfbeta}{{\bm{\beta}}}
\newcommand{\bfrho}{\bm{\rho}}
\newcommand{\bfep}{\bm{\epsilon}}
\newcommand{\bfga}{{\boldsymbol\gamma}}
\newcommand{\bfdelta}{\boldsymbol\delta}
\newcommand{\bfK}{\mathbf{K}}
\newcommand{\bfk}{\mathbf{k}}
\newcommand{\bfone}{\mathbf{1}} %bold face 1, indicator counting function
\newcommand{\bfzero}{\mathbf{0}}
\newcommand{\bfM}{\mathbf{M}}
\newcommand{\bbM}{\mathbb{M}}
\newcommand{\bbL}{\mathbb{L}}
\newcommand{\cX}{\mathcal{X}}
\newcommand{\cM}{\mathcal{M}}
\newcommand{\cI}{\mathcal{I}}
\newcommand{\cF}{\mathcal{F}}
\newcommand{\ccM}{\mathscr{M}}
\newcommand{\cJ}{\mathcal{J}}
\newcommand{\cE}{\mathcal{E}}
\newcommand{\cT}{\mathcal{T}}
\newcommand{\cS}{\mathcal{S}}
\newcommand{\rmH}{\mathrm{H}}
\newcommand{\rmI}{\mathrm{I}}
\newcommand{\Atil}{\Tilde{A}}
\newcommand{\Omegatil}{\Tilde{\Omega}}
\newcommand{\negj}{{\neg j}}
\newcommand{\negi}{{\neg i}}
\newcommand{\bfC}{\mathbf{C}}
\newcommand{\bsj}{{\backslash j}}
\newcommand{\bfv}{\mathbf{v}}
\newcommand{\bft}{\mathbf{t}}
\newcommand{\bfu}{\mathbf{u}}
\newcommand{\bfa}{\mathbf{a}}
\newcommand{\bfb}{\mathbf{b}}
\newcommand{\bfc}{\mathbf{c}}
\newcommand{\bfd}{\mathbf{d}}
\newcommand{\ci}{{(i)}}
\newcommand{\cj}{{(j)}}
\newcommand{\bfs}{\mathbf{s}}
\newcommand{\bfxi}{\boldsymbol\xi}
\newcommand{\kf}{{(k,1)}}
\newcommand{\ks}{{(k,2)}}
\newcommand{\kt}{{(k,3)}}
\newcommand{\ki}{{(k,i)}}
\newcommand{\supp}{\mathrm{supp}}
\newcommand{\bfzeta}{\boldsymbol\zeta}

% Math terms
\newcommand{\MLE}{\mathrm{MLE}}
\newcommand{\MAP}{\mathrm{MAP}}
\newcommand{\EB}{\mathrm{EB}}
\newcommand{\EM}{\mathrm{EM}}
\newcommand{\R}{\mathbb{R}}
\newcommand{\iid}{\mathrm{i.i.d.}}
\newcommand{\conv}{\mathrm{conv}} %convex hull
\newcommand{\mrf}{\mathrm{MRF}}
\newcommand{\bn}{\mathrm{BN}}
\newcommand{\fg}{\mathrm{FG}}
\newcommand{\ow}{\mathrm{otherwise}}
\newcommand{\ri}{\mathrm{ri}}
\newcommand{\ra}{\rightarrow}
\newcommand{\la}{\leftarrow}
\newcommand{\Cat}{\mathrm{Cat}}

% Graph Theory
\newcommand{\Pa}{\mathrm{Pa}}
\newcommand{\Ch}{\mathrm{Ch}}
\newcommand{\Nb}{\mathrm{Nb}}
\newcommand{\Anc}{\mathrm{Anc}}
\newcommand{\Desc}{\mathrm{Desc}}
\newcommand{\mc}{\mathrm{mc}}
\newcommand{\smc}{\mathrm{c}}
\newcommand{\cC}{\mathcal{C}} % cliques
\newcommand{\Para}{\mathrm{Para}}

\begin{abstract}
\ninept
In traditional multivariate data analysis, dimension reduction and regression have been treated as distinct endeavors. Established techniques such as principal component regression (PCR) and partial least squares (PLS) regression traditionally compute latent components as intermediary steps—although with different underlying criteria—before proceeding with the regression analysis. In this paper, we introduce an innovative regression methodology named PLS-integrated Lasso (PLS-Lasso) that integrates the concept of dimension reduction directly into the regression process. We present two distinct formulations for PLS-Lasso, denoted as PLS-Lasso-v1 and PLS-Lasso-v2, along with clear and effective algorithms that ensure convergence to global optima. PLS-Lasso-v1 and PLS-Lasso-v2 are compared with Lasso on the task of financial index tracking and show promising results. 
\end{abstract}
\begin{keywords}\ninept
Least absolute shrinkage and selection operator, Partial least squares regression, Statistical learning 
\end{keywords}

\section{Introduction}
\ninept
\label{sec:intro}
In contemporary statistical modeling and data analysis, regression and dimension reduction are fundamental tasks with far-reaching applications. Traditionally, dimension reduction precedes regression, offering noise reduction and enhanced visualization. Notable examples include principal component regression (PCR) \cite{pcr} and partial least squares (PLS) regression \cite{pls_wold}. While the former finds the directions of maximum variance in the input data, the latter seeks the directions that accounts for maximum covariance between the input and output data. 

In recent decades, there has been significant development within the family of least absolute shrinkage and selection operator (Lasso) variants \cite{tibshirani_lasso, sparsity_book}. Numerous new additions have emerged in this family, such as the elastic net \cite{elastic_net}, designed to enhance robustness in the presence of collinearities. Additionally, specialized techniques like group Lasso \cite{group_lasso} for handling feature grouping, and fused lasso \cite{fused_lasso} for dealing with naturally ordered data, have been introduced.

This paper introduces a novel approach called "PLS-integrated Lasso" (PLS-Lasso), integrating regularized regression and dimension reduction in a single framework. We present PLS-Lasso-v1 and PLS-Lasso-v2 formulations, each supported by an efficient algorithm that ensures global optimum convergence. PLS-Lasso-v1 blends disparate objectives with a new hyperparameter, while PLS-Lasso-v2 omits an extra hyperparameter. Through comparisons with Lasso in financial index tracking, we demonstrate the efficacy of PLS-Lasso. This method extends the Lasso family and opens doors to new possibilities.

This paper unfolds as follows: Section 2 presents the background for our approach. Section 3 introduces the existing sparse PLS (SPLS) method, and unveils PLS-Lasso models. Finally, section 4 gives experimental results in index tracking and section 5 concludes the article.

\section{Backgrounds}
\ninept
Before reviewing some regularized regression algorithms and multivariate analysis methods that are useful for developing the PLS-integrated LASSO, we introduce some notations. 
Consider a training dataset $\{(\bfx_i, y_i)\}_{i=1}^N$, with $\bfx_i \in \R^d$ and $y_i \in \R$. Let $X$ denote the $N \times d$ design matrix and $\bfy$ the $N$-dimensional response vector. We assume that $X$ and $\bfy$ are centralized. The $j$-th column of $X$ is represented as $X_j$, and the $i$-th entry of the $d$-dimensional weight vector $\bfw$ is denoted $w_i$.

\subsection{Least Absolute Shrinkage and Selection Operator (Lasso)}
The constrained formulation of the Lasso is given by \cite{tibshirani_lasso}:
\begin{equation}
\begin{aligned}
    \underset{\bfw}{\text{minimize}}\quad& \frac{1}{2}\|X\bfw -\bfy\|^2 \\
    \text{subject to}\quad& \|\bfw\|_1 \leq \epsilon,
\end{aligned}
\label{eq:lassoConst}
\end{equation}
where $\|\bfw\|_1$ denotes the $L_1$ norm of $\bfw$, and $\epsilon$ is a positive constant that controls the level of sparsity in the solution. The Lagrangian form of Lasso is given by \cite{tibshirani_lasso}:
\begin{equation}
\begin{aligned}
    \underset{\bfw}{\text{minimize}}\quad& \frac{1}{2}\|X\bfw -\bfy\|^2 + \lambda \|\bfw\|_1.
\end{aligned}
\label{eq:lassoLag}
\end{equation}
Including an $L_1$ penalty term in the objective function motivates each entry of $\bfw$ to shrink towards zero \cite{hastie_esl}, thereby promoting sparsity. It has been established that under specific conditions, there exists a one-to-one correspondence between the constrained form~\eqref{eq:lassoConst} and the Lagrangian form~\eqref{eq:lassoLag} of Lasso \cite{boyd_cvx}. Various algorithms exist for efficiently solving the Lasso problem, one notable example being the Iterative Shrinkage-Thresholding Algorithm (ISTA) \cite{beck_book,ista}.

The $L_1$ penalty term can be generalized to a penalty function $\phi:\mathbb{R}^n \rightarrow \mathbb{R}_+$ that is concave over the nonnegative orthant \cite{zhao2006model}. Examples of such penalty functions include \cite{palomar}:
\begin{itemize}
    \item $L_0$ norm,
    \item $L_p$ norm, with $0<p<1$,
    \item $\phi(\bfw) = d - \sum_{i=1}^d \exp(-|w_i|\slash p)$, with $p>0$,
    \item $\phi(\bfw) = \frac{1}{\log(1+1\slash p)}\sum_{i=1}^d \log(1+|w_i|\slash p)$, with $p>0$.
\end{itemize}
%the $L_0$ norm, ${1 - exp(-|x|\slash p)}$, ${\frac{log(1+|x|\slash p)}{log(1+1\slash p)}}$, as illustrated in figure \ref{fig:pen_fun} \cite{palomar}.  
%\begin{figure}[htb]
%    \centering
%    \includegraphics[width=1\linewidth]{l1plot.png}
%    \caption{Graphs of various penalty functions $\phi$.}
%    \label{fig:pen_fun}
%\end{figure}

\subsection{One-step Thresholding (OST)}
One-Step Thresholding (OST) is an algorithm that solves regression problems subject to a cardinality constraint. Formally, the problem can be described as:
\begin{equation}
\begin{aligned}
\underset{\bfw}{\text{minimize}}\quad & \frac{1}{2}\|X\bfw - \bfy\|^2\\
\text{subject to}\quad & \|\bfw\|_0 = K,
\end{aligned}
\label{eq:ost}
\end{equation}
where $\|\bfw\|_0$ denotes the $L_0$ norm of $\bfw$, and $K$ is the sparsity level. The algorithm evaluates $\bfz = X^\intercal \bfy$ to obtain the correlation coefficients between each feature column of $X$ and the response vector $\bfy$. It then selects the indices corresponding to the $K$ largest magnitudes in $\bfz$. OST has proven to be effective under certain conditions \cite{wright_ma,ost_tropp, ost_cai, ost_davenport}. 
%Specifically, if $\bfw$ is $K$-sparse and $X\bfw = \bfy$, OST recovers the support (i.e. the index set of its nonzero entries) of $\bfw$ provided that
%$$\frac{\min_{i} |w_i|}{\|\bfw\|_1} \geq \frac{2\mu(X)}{\mu(X)+1},$$
%where $\mu(X)$ represents the incoherence of $X$, defined as
%$$\mu(X) = \max_{i\neq j} \left| \left\langle \frac{X_i}{\|X_i\|}, \frac{X_j}{\|X_j\|}\right\rangle \right|.$$

\subsection{Partial Least Squares (PLS) Regression}
Introduced by \cite{pls_wold} through the nonlinear iterative partial least squares (NIPALS) algorithm, partial least squares (PLS) regression finds the directions of maximal covariance between the input and output data in a sequential manner. 
In the context of PLS, latent components are constructed as a weighted sum of the original features.
The weights are often referred to as directions as they define a subspace for which the latent components span. 
Here, we focus on the special case of PLS where the response variable $y$ is univariate, and adopt the probabilistic interpretation given in \cite{pls_prob}.
The first latent factor $t_1 = \bfw_1^\intercal \bfx$ is found by solving
\begin{equation}
\begin{aligned}
    \underset{\bfw_1}{\text{maximize}}\quad &\big\{\mathrm{Cov}[\bfw_1^\intercal \bfx, y] = \bfw_1^\intercal \Sigma_{\bfx y}\Sigma_{y\bfx} \bfw_1\big\}\\
    \text{subject to}\quad &\|\bfw_1\|_2 = 1. 
\end{aligned}
\label{eq:pls1st}
\end{equation}
The covariance term $\Sigma_{\bfx y}$ can be approximated by the empirical covariance that is proportional to $X^\intercal y$. Thus the optimization problem~\eqref{eq:pls1st} becomes
\begin{equation}
\begin{aligned}
    \underset{\bfw_1}{\text{maximize}}\quad &(\bfy^\intercal X \bfw_1)^2\\
    \text{subject to}\quad &\|\bfw_1\|_2 = 1. 
\end{aligned}
\end{equation}
The subsequent weight vectors $\bfw_k$ are derived by solving
\begin{equation}
\begin{aligned}
    \underset{\bfw_k}{\text{maximize}}\quad &(\bfy^\intercal X \bfw_k)^2\\
    \text{subject to}\quad &\bfw_k^\intercal X^\intercal X\bfw_j = 0,\, \forall j<k\\
    &\|\bfw_k\|_2 = 1.
\end{aligned}
\label{eq:pls2nd}
\end{equation}
The first constraint of~\eqref{eq:pls2nd} ensures that each new latent variable $t_k = \bfw_k^\intercal \bfx$ is uncorrelated to its predecessors.

\subsection{Multi-objective Optimization}
Multi-objective optimization problems are typically formulated as \cite{moo_deb}
\begin{equation}
\begin{aligned}
\underset{x \in \mathcal{X}}{\text{minimize}} \quad& \big(f_1(x), f_2(x),\dots, f_K(x)\big),
\end{aligned}
\end{equation}
where each $f_i:\mathcal{X} \rightarrow \mathbb{R}$. A simple way of solving such problems is linear scalarization \cite{moo_vec}, where the objective function is transformed into a weighted sum of the component functions as demonstrated below
\begin{equation}
\begin{aligned}
\underset{x \in \mathcal{X}}{\text{minimize}}\quad& w_1 f_1(x) + w_2 f_2(x) + \dots + w_K f_K(x), 
\end{aligned}
\end{equation}
where $w_1, \dots, w_K \geq 0$. The weights are assigned to each component function according to their relative importance. Despite being straightforward and intuitive, this method lacks a sense of rigor due to the arbitrary weight assignment. 

\section{PLS-integrated Lasso}
\ninept
In this section, we first review the sparse PLS method, and then present our PLS-integrated Lasso methods.

\subsection{Sparse Partial Least Squares (SPLS) Regression}
The sparse partial least squares (SPLS) regression is sparse in the sense that each latent component $t_k$ is a sparse linear combination of the original features $\bfx$. 
Suppose that there are $K$ latent factors being extracted, and the weight vector $\bfw_k$ is sparse with support (i.e. the index set of its nonzero entries) $\supp(\bfw_k)$; then the SPLS solution is sparse with support $\bigcup_{k=1}^K \supp(\bfw_k)$. 

Unlike traditional PLS, SPLS induces sparsity in the first direction by imposing an $L_1$ penalty term, as given by the following optimization problem:
\begin{equation}
\begin{aligned}
    \underset{\bfw}{\text{maximize}}\quad &(\bfy^\intercal X \bfw)^2 - \lambda \|\bfw\|_1\\
    \text{subject to}\quad& \|\bfw\|_2 = 1.
\end{aligned}
\end{equation}
To further promoting sparsity, a generalized SPLS formulation has been proposed \cite{spls_chun}:
\begin{equation}
\begin{aligned}
    \underset{\bfw,\bfc}{\text{minimize}}\quad &-k (\bfy^\intercal X \bfw)^2 + (1-k) ((\bfc-\bfw)^\intercal X^\intercal \bfy)^2\\
    &\indent +\lambda_1 \|\bfc\|_1 + \frac{\lambda_2}{2}\|\bfc\|^2 \\
    \text{subject to}\quad& \|\bfw\|_2 = 1,
\end{aligned}
\end{equation}
where $k \in (0,1]$. The optimal vector $\bfc$ is rescaled to unit norm and used as the latent direction. Subsequent latent directions are extracted in a manner analogous to ~\eqref{eq:pls2nd}. Notably, most existing SPLS algorithms employ two-step approaches, as PLS inherently performs both dimension reduction and regression.

\subsection{PLS-Lasso-v1}
The inception of the PLS-integrated Lasso (PLS-Lasso) method draws inspiration from both the OST algorithm and the SPLS method. These methods share a commonality: SPLS computes the initial latent factor by maximizing its covariance with the response variable, as encapsulated by $(\bfy^\intercal X \bfw)^2$, whereas the OST algorithm employs the correlation coefficient between $\bfy$ and $ X \bfw$, as its guiding criterion.

The essence of PLS-Lasso lies in the parallel pursuit of regression and dimension reduction. While traditional approaches, such as principal component regression (PCR), often involve performing dimension reduction first and then conducting regression on the extracted latent factors, PLS-Lasso offers a unified methodology where these tasks are not isolated. The task of regression aims to minimize the sum of squared residuals, while the task of dimension reduction under the PLS settings seeks the direction that maximizes the covariance between the input and output data. The conventional divergence between the two objectives can be addressed through a straightforward multi-objective optimization approach, specifically via scalarization. The PLS-Lasso-v1 is formulated as follows:
\begin{equation}
\begin{aligned}
    \underset{\bfw}{\text{minimize}}\quad &\frac{1}{2}\|X\bfw -\bfy\|^2 - \frac{\mu}{2}(\bfy^\intercal X \bfw)^2\\
    \text{subject to}\quad &\|\bfw\|_1 \leq \epsilon,
\end{aligned}
\label{eq:pls_lasso_v1_1}
\end{equation}
and its Lagrangian form is given by:
\begin{equation}
\begin{aligned}
    \underset{\bfw}{\text{minimize}}\quad \frac{1}{2}\|X\bfw -\bfy\|^2 - \frac{\mu}{2}(\bfy^\intercal X \bfw)^2 + \lambda \|\bfw\|_1,
\end{aligned}
\label{eq:pls_lasso_v1_2}
\end{equation}
where $\mu>0$. Specially, if $\mu$ is set to zero, PLS-Lasso-v1 reduces to the Lasso approach. Note that the formulation ~\eqref{eq:pls_lasso_v1_2} is equivalent to
\begin{equation}
\begin{aligned}
    \underset{\bfw}{\text{minimize}} \quad \frac{1}{2} \bfw^\intercal X^\intercal (I-\mu \bfy \bfy^\intercal)X\bfw  -\bfy^\intercal X\bfw + \lambda \|\bfw\|_1.
\end{aligned}
\label{eq:pls_lasso_v1_3}
\end{equation}
The optimization problem ~\eqref{eq:pls_lasso_v1_3} is convex as long as $0 \leq \mu \leq {1 \slash\|\bfy\|^2}$, making it easily solvable with convex optimization solver like CVX \cite{cvx_matlab}. This paper provides two algorithms for solving~\eqref{eq:pls_lasso_v1_3}, and given the convexity of the formulation, both algorithms should converge to a global optimum. 
\begin{defn} Given a positive number $\lambda$, the soft-thresholding function for $w \in \R$ is defined as
$$\mathcal{S}_\lambda(w) = [|w|-\lambda]_+ \mathrm{sgn}(w).$$
The soft-thresholding function for vector $\bfw \in \R^d$ is defined as
$$\mathcal{S}_\lambda(\bfw) = [\mathcal{S}_\lambda(w_i)]_{i=1}^d.$$
\end{defn}
The optimization~\eqref{eq:pls_lasso_v1_3} solved by the iterative soft-thresholding algorithm (ISTA)\cite{ista} is outlined in algorithm 1. Alternatively, the Douglas-Rachford method, presented in Algorithm 2, offers another approach for solving this problem. Algorithm 2, belonging to the genre of distributed optimization algorithms, ensures rapid convergence.

\begin{algorithm}
\caption{PLS-Lasso-v1 (ISTA)}%\label{algorithm}
\begin{small}
\KwIn{$t>0,\, \mu \in (0,1\slash \|\bfy\|^2],\, \lambda >0$}
Initialize $\bfw_0 = (X^\intercal X)^{-1} X^\intercal \bfy$\;
\Repeat{convergence}{
$\bfxi_k = \bfw_k - t\big(X^\intercal (I-\mu \bfy \bfy^\intercal)X\bfw_k - X^\intercal \bfy\big)$\;
$\bfw_{k+1} = \mathcal{S}_{\lambda t}(\bfxi_k)$\;
}
\KwOut{$\bfw_* \leftarrow \bfw_k$}
\end{small}
\label{alg:pls_lasso_v1a}
\end{algorithm}
Although PLS-Lasso-v1 is intuitive and easy to implement, simultaneously tuning the parameters $\mu$ and $\lambda$ may be a challenging endeavor. Moreover, identifying an optimal $\mu$ that guides the solution towards Pareto optimality within the feasible set can be difficult. 
\begin{algorithm}
\caption{PLS-Lasso-v1 (Douglas-Rachford)}%\label{algorithm}
\begin{small}
\KwIn{$\mu \in (0,1\slash \|\bfy\|^2]$,\, $ \lambda >0$,\, and $\rho = 1$}
Initialize $\bfw_0 = (X^\intercal X)^{-1} X^\intercal \bfy$,\, $\bfxi_0 = \bfw_0$,\, and $\bfzeta_0 = \bfw_0$\;
\Repeat{convergence}{
$\bfw_{k+1} = (X^\intercal(I-\mu \bfy \bfy^\intercal)X + \rho I)^{-1} (X^\intercal \bfy + \rho \bfxi_k + \rho \bfzeta_k)$\;
$\bfxi_{k+1} = \mathcal{S}_{\lambda \slash \rho}(\bfw_{k+1} - \bfzeta_k)$\;
$\bfzeta_{k+1} = \bfzeta_k + \rho(\bfxi_{k+1} - \bfw_{k+1})$\;
}
\KwOut{$\bfw_* \leftarrow \bfw_k$}
\end{small}
\label{alg:pls_lasso_v1b}
\end{algorithm}
\vspace{-10pt}

\subsection{PLS-Lasso-v2}
The PLS-Lasso-v2 formulation eliminates the need for an additional hyperparameter $\mu$. Consider the optimization problem:
\begin{equation}
\begin{aligned}
    \underset{\bfw}{\text{minimize}}\quad & \left\{ f(\bfw) := \frac{1}{2}\left\|\frac{X\bfw-\bfy}{\bfy^\intercal X\bfw}\right\|^2\right\}\\
    \text{subject to}\quad &\|\bfw\|_1 \leq \epsilon,
\end{aligned}
\label{eq:pls_lasso_v2_1}
\end{equation}
and its equivalent Lagrangian form:
\begin{equation}
\begin{aligned}
    \underset{\bfw}{\text{minimize}} \quad f(\bfw) + \lambda \|\bfw\|_1. 
\end{aligned}
\label{eq:pls_lasso_v2_2}
\end{equation}
The objective function $f$ is quasi-convex, a result of composing a squared norm with a linear fractional transformation \cite{boyd_cvx}.
%as it can be regarded as a composition of a convex squared norm function with a linear fractional transformation, which is quasi-convex \cite{boyd_cvx}. 
Importantly, the optimal solution $\bfw^*$ that solves \eqref{eq:pls_lasso_v2_1} guarantees a positive $\bfy^\top X\bfw^*$. For demonstration, we express $f$ as:
\begin{equation}
\begin{aligned}
    f(\bfw) &= \frac{\|X\bfw\|^2 + \|\bfy\|^2}{2\|\bfy^\intercal X\bfw\|^2} - \frac{1}{\bfy^\intercal X\bfw},
\end{aligned}
\label{eq:pls_lasso_v2_alt}
\end{equation}
from which it follows that if $\bfy^\top X\bfw^* < 0$, a contradiction arises: $f(-\bfw^*) < f(\bfw^*)$. We make the following Charnes-Cooper transformation \cite{cctrans}:
\begin{equation}
\begin{aligned}
    \bfga = \frac{\bfw}{\bfy^\intercal X\bfw},\, t = \frac{1}{\bfy^\intercal X\bfw}.
\end{aligned}
\label{eq:pls_lasso_v2_3}
\end{equation}
This leads to the reformulated optimization problem~\eqref{eq:pls_lasso_v2_4}:
\begin{equation}
\begin{aligned}
    \underset{\bfga,t}{\text{minimize}}\quad &\frac{1}{2}\|X\bfga -t\bfy\|^2 + \frac{\lambda}{t}\|\bfga\|_1\\
    \text{subject to}\quad & \bfy^\intercal X\bfga = 1\\
    & t>0.
\end{aligned}
\label{eq:pls_lasso_v2_4}
\end{equation}

We propose an iterative scheme based on the structure of~\eqref{eq:pls_lasso_v2_4}. In the $l$-th iteration, we fix $t_l$ and update $\bfga_{l+1}$ as the optimal solution to the following optimization problem:
\begin{equation}
\begin{aligned}
    \underset{\bfga}{\text{minimize}}\quad &\frac{1}{2} \bfga^\intercal X^\intercal X\bfga + \frac{\lambda}{t_l}\|\bfga\|_1 \\
    \text{subject to}\quad &\bfy^\intercal X\bfga = 1.
\end{aligned}
\label{eq:pls_lasso_v2_5}
\end{equation}
We employ an augmented Lagrangian formulation to transition from ~\eqref{eq:pls_lasso_v2_5} to~\eqref{eq:pls_lasso_v2_6}:
\begin{equation}
\begin{aligned}
    \underset{\bfga,\bfbeta}{\text{minimize}}\quad& \frac{1}{2}\bfga^\intercal X^\intercal X \bfga + \frac{\lambda}{t}\|\bfbeta\|_1 \\
    &\indent + \frac{\rho}{2}\|\bfga-\bfbeta\|^2 + \frac{\rho}{2} (\bfy^\intercal X\bfga - 1)^2 \\
    \text{subject to}\quad &\bfy^\intercal X\bfga = 1.
\end{aligned}
\label{eq:pls_lasso_v2_6}
\end{equation}
Algorithm 3, which utilizes the alternating direction method of multipliers (ADMM), solves the optimization problem~\eqref{eq:pls_lasso_v2_6}.

\begin{algorithm}
\caption{The Optimization using ADMM}%\label{algorithm}
\begin{small}
\KwIn{$\lambda$,\, $t_l$,\, $\bfga_0$,\, and $\rho = 1$}
Initialize $\bfbeta_0 = \bfga_0$,\, and $\nu_0 = 1$\;
\Repeat{convergence}{
$\bfga_{k+1} = (X^\intercal(I+\rho \bfy \bfy^\intercal)X +\rho I)^{-1}((\nu_{k}-\rho)X^\intercal \bfy - \rho \bfbeta_k)$\;
$\bfbeta_{k+1} = \mathcal{S}_{\frac{\lambda}{\rho t_l}}(\bfga_{k+1})$\;
$\nu_{k+1} = \nu_k + \rho(\bfy^\intercal X \bfga_{k+1} -1)$\;
}
\KwOut{$\bfga_* \leftarrow \bfga_k$}
\end{small}
\label{alg:pls_lasso_v2a}
\end{algorithm}
After obtaining $\bfga_{l+1}$ from Algorithm 3, $t_{l+1}$ can be updated as the optimal solution to the following univariate optimization problem:
\begin{equation}
\begin{aligned}
    \underset{t}{\text{minimize}}\quad &\frac{\|\bfy\|^2}{2}t^2 -t  + \frac{\lambda \|\bfga_{l+1}\|_1}{t}\\
    \text{subject to}\quad &t>0.
\end{aligned}
\label{eq:pls_lasso_v2_7}
\end{equation}
Finally, the iterative scheme for solving PLS-Lasso-v2 is summarized in Algorithm 5. Given the quasi-convex nature of formulation~\eqref{eq:pls_lasso_v2_6}, the proposed iterative scheme ensures convergence to a global minimum.

\begin{algorithm}
\caption{PLS-Lasso-v2 (Iterative Scheme)}%\label{algorithm}
\begin{small}
\KwIn{$\lambda >0$,\, and $\rho = 1$}
Initialize $\bfw_0 = (X^\intercal X)^{-1} X^\intercal \bfy$,\, $t_0 = 1\slash \bfy^\intercal X\bfw_0$,\, $\bfga_0 = t_0\bfw_0$,\,and $\bfbeta_0 = \bfga_0$\;
\Repeat{convergence}{
Given $\bfga_l$ and $t_l$, set $\bfga_{l+1}$ as the output of the algorithm $3$\;
Given $\bfga_{l+1}$, calculate the  optimal solution to the univariate optimization problem $(18)$ through bisection and set $t_{l+1}$ equal to its optimal solution\;
}
\KwOut{$\bfw_* \leftarrow \bfga_l \slash t_l$}
\end{small}
\label{alg:pls_lasso_v2b}
\end{algorithm}

% Boyd CVX book P97
% https://iwct.sjtu.edu.cn/personal/yingcui/Slides/CO/3-convex%20functions.pdf P37
% my note
% (12) objective function 
%       linear fractional
%       lemma: quasiconvex
%       lemma: surrogate function
% lemma: MM algorithm converges to a global minimum

\section{Financial Index Tracking}              
% frontier graph                                
% nested cross validation, explain the procedure
% should i use a table? how do i do it?   
In this section, we use the NASDAQ 100 index and the S\&P 500 index datasets, collected from the time interval between November 2004 and April 2016, to demonstrate the efficacy of our proposed methods\cite{fin_data}. The SP500 dataset encompasses 595 weekly stock returns for 442 distinct companies, in addition to the S\&P 500 index. The NASDAQ100 dataset incorporates weekly stock returns for 82 companies and the corresponding weekly NASDAQ 100 index returns spanning 596 weeks. 
Our primary objective is to uncover an optimal sparse linear combination of stock returns that can effectively predict market index returns.
To evaluate the predictive performance, we employ the root mean square error (RMSE) as the criterion. Note that we relax the sum-to-one and non-negativity constraints on asset weights to solely focus on the predictive accuracy of our new Lasso variant in regression. 

For both the NASDAQ 100 and S\&P 500 datasets, we partition the data into two segments: the first 400 data points are used for training, while the remaining ones are used for testing the predictive power of the models. The key outcomes of our experiments are visualized in Figure 1, illustrating the relationship between cardinality and both training and testing errors across a range of hyperparameters ($\lambda$ values).

As depicted in Figure 1, PLS-Lasso-v2 consistently demonstrates Pareto optimality over Lasso in terms of lower testing and training errors and higher sparsity. Meanwhile, the performance of PLS-Lasso-v1 exhibits a certain degree of instability; in scenarios where the sparsity level is relatively high, PLS-Lasso-v1 outperforms its counterparts, but as sparsity decreases, both its training and testing errors increase. It is evident that PLS-Lasso-v1 requires meticulous tuning of hyperparameters for optimal performance.
%PLS-Lasso-v1 should be employed with discretion and an ad-hoc approach in mind.

%nested cross validation
%The NASDAQ 100 dataset contains the daily stock returns for 101 companies and the daily NASDAQ 100 index returns over a period of 399 days, from which the first 300 data points are used for training and the remaining ones are used for testing. The S\&P 500 data set contains 595 daily stock returns for 442 companies as well as the S\&P index. We use the first 400 days for training and the remaining 195 days for testing. 
\begin{figure}[htb]
    \centering
    \includegraphics[width=\linewidth]{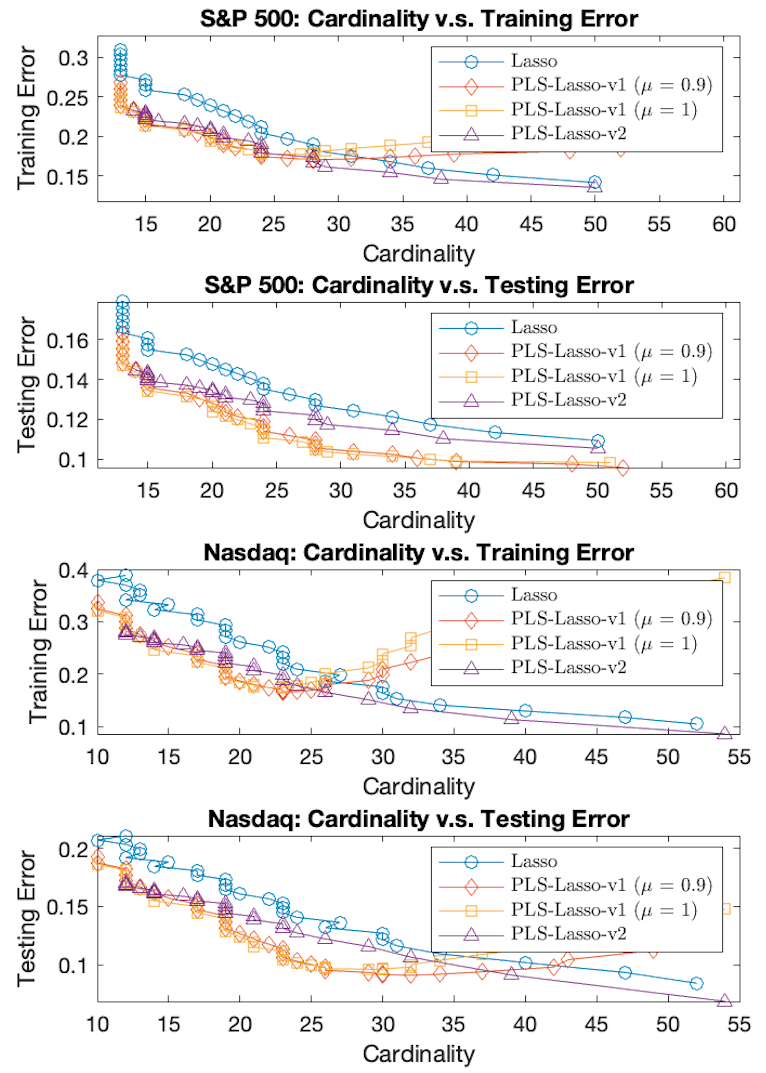}
    \caption{Comparison between Lasso, PLS-Lasso-v1 and PLS-Lasso-v2 on NASDAQ 100 and S\&P 500 stock returns data\vspace{-10pt}}
    \label{fig:my_label}
\end{figure}
%We then run multiple experiments with different $\lambda$ and the results are shown in figure $2$. As can be seen, PLS-Lasso-v1 is Pareto optimal to Lasso as the blue points representing the experiments with the PLS-Lasso approach are generally on the lower left side of the red points representing the experiments done by Lasso. 

\section{Conclusion and Future Work}
\ninept
In this paper, we introduced the methodology of PLS-integrated Lasso. As a new addition to the family of Lasso variant, PLS-Lasso unclocks new analytical prospects. We conclude the article by pointing out some limitations of PLS-Lasso and outline potential avenues for futere research:
\begin{enumerate}
\item PLS-Lasso currently addresses univariate response variables. Extending its application to encompass multivariate response variables would broaden its scope and applicability.
\item The present editions of PLS-Lasso focus on extracting the first latent direction. Further research can explore techniques to overcome the iterative limitations of the PLS algorithm, enabling the extraction of multiple latent components.
\item As most Lasso variants have a corresponding probabilistic counterparts, PLS-Lasso lacks a probabilistic interpretation. Establishing such connection could deepen its theoretical underpinnings. 
\item As described in Section 2.1, the $L_1$ regularization term can be substituted with generalized penalty functions. The resulting formulations are solvable by maximization-majorization algorithm as described in \cite{palomar,mm}. 
\end{enumerate}

\section{Acknowledgements}
\ninept
This research work is supported by a Math and Application Project (2021YFA1003504) under the National Key R\&D Program, a Collaborative Research Fund by RGC of Hong Kong (Project No. C1143-20G), a grant from the Natural Science Foundation of China (U20A20189), a grant from ITF - Guangdong-Hong Kong Technology Cooperation Funding Scheme (Project Ref. No. GHP/145/20), and an InnoHK initiative of The Government of the HKSAR for the Laboratory for AI-Powered Financial Technologies.
\label{sec:refs}
\bibliographystyle{IEEEbib} 
\bibliography{refs}

\end{document}